\pdfoutput=1

\documentclass[11pt]{article}

\usepackage[]{acl2023}

\usepackage{times}
\usepackage{latexsym}

\usepackage[T1]{fontenc}

\usepackage[utf8]{inputenc}

\usepackage{microtype}

\usepackage{inconsolata}

\usepackage{amsmath}
\usepackage{amssymb}
\usepackage{booktabs}
\usepackage{graphicx}
\usepackage{subcaption}
\usepackage{xcolor,colortbl}
\usepackage{comment}

\newcommand{\ours}{\textsf{MCKD}}
\newcommand{\oursfull}{Multistage Collaborative Knowledge Distillation from an LLM}
\usepackage{url}
\usepackage{enumitem}

\newcommand{\done}{\cellcolor{black}done} 
\newcommand{\zjc}[1]{{\textcolor{blue}{J:#1}}}

\newcommand{\eg}{\textit{e.g.}}
\newcommand{\ie}{\textit{i.e.}}

\newlist{myitemize}{description}{10}
\setlist[myitemize]{labelindent=4pt, leftmargin=0pt, align=left, noitemsep, topsep=0pt}

\title{
Multistage Collaborative Knowledge Distillation from a Large Language Model for Semi-Supervised Sequence Generation
}

\author{
Jiachen Zhao$^1$, Wenlong Zhao$^{1}$\thanks{~~Equal contribution}~, Andrew Drozdov$^{1}$\footnotemark[1]~, Benjamin Rozonoyer$^1$, \\
\textbf{Md Arafat Sultan$^2$, Jay-Yoon Lee$^3$, Mohit Iyyer$^1$, Andrew McCallum$^1$} \\
$^1$University of Massachusetts Amherst,~$^2$IBM Research AI,~$^3$Seoul National University\\
\texttt{\{jiachenzhao,wenlongzhao,adrozdov\}@umass.edu}\\
}

\begin{document}

\maketitle

\begin{abstract}
We study semi-supervised sequence generation tasks, where the few labeled examples are too scarce to finetune a model, and meanwhile, few-shot prompted large language models (LLMs) exhibit room for improvement.
In this paper, we present the discovery that a student model distilled from a few-shot prompted LLM can commonly generalize better than its teacher to unseen examples on such tasks. 
We find that the student is able to learn a general pattern from the high-quality pseudolabels produced by the teacher during knowledge distillation (KD), and favorably not a general pattern from the low-quality pseudolabels.
Leveraging this discovery, we propose a new method, \oursfull~(\ours), for these tasks.
\ours{} first few-shot prompts an LLM to produce pseudolabels for unlabeled data. 
Then at each stage of an iterative KD process, a new pair of students is trained on disjoint partitions of the pseudolabeled data, and produces new and improved pseudolabels for their unseen partitions.
We conduct extensive experiments on four syntactic and semantic parsing datasets and show the effectiveness of \ours\ for low-resource semi-supervised sequence generation.
On CRAFT biomedical parsing, for example, 3-stage \ours{} with 50 labeled examples outperforms an LLM teacher and vanilla KD by 7.5\% and 3.7\% parsing F1, respectively, and matches the performance of supervised finetuning with 500 labeled examples.~\footnote{Our code is made available \href{https://github.com/andotalao24/Multistage-Collaborative-Knowledge-Distillation}{github.com/andotalao24/Multistage-Collaborative-Knowledge-Distillation}.}
\end{abstract}

\section{Introduction}


Low-resource tasks are common in real life, including within specialized domains, since data annotation often requires expert knowledge and incurs significant costs~\citep{verspoor2012corpus}.
Semi-supervised learning has been proposed as a solution when abundant unlabeled data are available~\citep{Blum1998CombiningLA,DBLP:conf/naacl/McCloskyCJ06,han2018co,kontonis2023slam}. 
In a typical application, a model is trained on limited labeled data and produces pseudolabels for unlabeled data~\citep{DBLP:conf/coling/McCloskyCJ08,amini2022self}. 
The pseudolabeled data are then filtered according to confidence thresholds and used to train a new model. 
In more extreme few-shot cases, labeled data are too scarce to finetune a model to begin with. 
Large language models (LLMs) offer a useful mechanism for synthesizing pseudolabels, thanks to their remarkable ability to learn \textit{in context} from only a handful of demonstrations ~\citep{wang-etal-2021-want-reduce,yoo-etal-2021-gpt3mix-leveraging,ding2022gpt,wang-etal-2023-lets}.
Smaller and faster models can then be trained for these tasks using knowledge distillation (KD) from LLMs.

\looseness=-1
In this paper, we study a challenging semi-supervised sequence generation setting where labeled data are too few to finetune a model and few-shot prompted LLMs exhibit room for improvement.
These will happen when the task is both expensive to annotate and under-represented in the pretraining of off-the-shelf LLMs. 
For example, it took 80 annotators around 2.5 years to parse 20k sentences of biomedical text in the CRAFT corpus~\cite{verspoor2012corpus}. Meanwhile, pretrained LLMs do not always excel at tasks in specialized domains \cite{Kung2022,Singhal2022LargeLM} and tasks that involve specialized structured outputs, \eg{}, semantic or syntactic parsing.
The overarching research question of this paper is whether LLMs can still be leveraged in such scenarios to develop strong prediction models.

To this end, we examine knowledge distillation (KD) from a few-shot prompted LLM to a much smaller model. We discover that the student can commonly outperform its LLM teacher on \textbf{unseen evaluation data} from such tasks.
Our analysis reveals that the student can learn a general pattern from high-quality pseudolabels from the teacher, while helpfully failing to capture a general pattern for the low-quality ones due to their noisy nature.
This discovery is encouraging because it opens up the possibility for leveraging the KD students as teachers for further distillation.


\looseness=-1
Leveraging the discovery, we propose a novel method, Multistage Collaborative KD from an LLM~(\ours), for semi-supervised sequence generation.
\ours{} first collects pseudolabels for a large amount of unlabeled data from a few-shot prompted LLM, bootstraping a multistage KD process.
At each KD stage, a new pair of students is trained on distinct partitions of the pseudolabeled data and asked to produce pseudolabels for the data that they have not been trained on.
This process improves upon vanilla KD through two core mechanisms: (1) cross-partition labeling at each KD stage, which splits all pseudolabeled data into two mutually exclusive partitions to train a \textbf{collaborative} pair of students and leverages their strong generalization capabilities to relabel the data, and (2) \textbf{multistage} KD, whereby student models continue to generalize better and produce higher quality pseudolabels than their teachers over multiple KD stages.

\ours\ outperforms the LLM teacher and the relevant KD baselines in our evaluation, and is competitive with supervised finetuning with many more labeled data.
On CRAFT constituency parsing, for example, \ours{} with 50 labeled examples outperforms the LLM and vanilla KD by 7.5\% and 3.7\% F1, respectively, and matches supervised finetuning with 500 examples. 
On ATIS semantic parsing, \ours{} with 50 labeled examples outperforms the LLM and vanilla KD by 7.2\% and 2.3\% F1, matching supervised finetuning with $>$750 examples.
The following is a summary of our contributions:
\begin{itemize}[leftmargin=10pt,topsep=0pt,noitemsep]
    \item We study if LLMs can be leveraged to train fast and accurate student models for semi-supervised sequence generation tasks where the LLM itself exhibits room for improvement when prompted with in-context learning examples. We find that KD students can often achieve better generalization than their LLM teachers (Section \ref{sec:fidelity}).
    \item We propose \ours{}, a novel KD-based solution for such tasks. Data partitioning and cross-partition labeling enable \ours{} to gradually improve the quality of pseudolabels over multiple stages of distillation, yielding increasingly better generations of students (Section \ref{sec:mistkd}).
    \item \ours{} substantially outperforms prompted LLM, finetuned, and KD baselines on multiple low-resource tasks (Section \ref{sec:setup}-\ref{sec:results}).
    \item Further analyses show that: (\textit{a})~\ours{} students correct many of the teacher's errors, and (\textit{b})~\ours{} scales well with increasing amounts of available unlabeled data (Section \ref{sec:correct}-\ref{sec:size}).
\end{itemize}

\section{\ours: \oursfull}\label{sec:mistkd}
\subsection{Problem setup}
Semi-supervised sequence-to-sequence generation tasks consist of few labeled examples $\mathcal{D}^{\text{labeled}}=\{x_i,y_i\}_{i=1}^{N_\text{labeled}}$ and many unlabeled examples $\mathcal{D}^{\text{unlabeled}}=\{x_i\}_{i=1}^{N_\text{unlabeled}}$ for training, where $x_i$ and $y_i$ are all sequences.
We focus on a scenario in which (1) $\mathcal{D}^{\text{labeled}}$ is not large enough for training a capable prediction network via direct supervised finetuning, and (2) LLM few-shot prompting with demonstrations sampled from $\mathcal{D}^{\text{labeled}}$ exhibits room for improvement.
This is a challenging but prevalent scenario that occurs when a task, such as parsing, is both expensive to annotate and under-represented in LLM pretraining.

\begin{figure*}[!ht]
    \centering
    \includegraphics[width=0.85\linewidth]{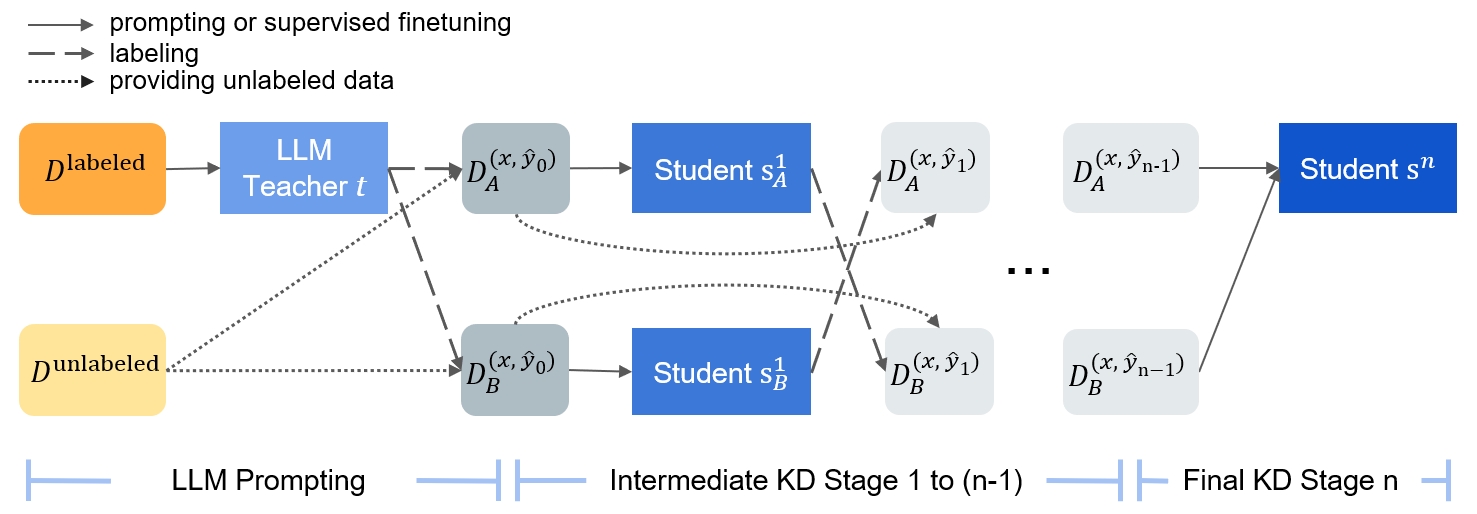}
    \caption{Overview of \ours. 
    \textbf{(1)} We use demonstrations from labeled data $\mathcal{D}^\text{labeled}$ to few-shot prompt an LLM teacher $t$ to produce pseudolabels for unlabeled data $\mathcal{D}^\text{unlabeled}$. 
    We partition $\mathcal{D}^\text{unlabeled}$ into $\mathcal{D}^\text{unlabeled}_A$ and $\mathcal{D}^\text{unlabeled}_B$, and let $\mathcal{D}^{(x, \hat{y}_0)}_A$ and $\mathcal{D}^{(x, \hat{y}_0)}_B$ denote the same partitions but with teacher-generated pseudolabels. 
    \textbf{(2)} At the $i$-th intermediate KD stage, students $s_A^i$ and $s_B^i$ are trained on previously pseudolabeled data $\mathcal{D}_A^{(x,\hat{y}_{i-1})}$ and $\mathcal{D}_B^{(x,\hat{y}_{i-1})}$, respectively, and leveraged to label the other partitions $\mathcal{D}_B^\text{unlabeled}$ and $\mathcal{D}_A^\text{unlabeled}$ and produce $\mathcal{D}_B^{(x,\hat{y}_i)}$ and $\mathcal{D}_A^{(x,\hat{y}_i)}$, which will be used to train the next-stage student(s).
    \textbf{(3)} In the final KD stage, a single final student $s^n$ is trained on both latest pseudolabeled partitions $\mathcal{D}^{(x, \hat{y}_{n-1})}_A$ and $\mathcal{D}^{(x, \hat{y}_{n-1})}_B$.}\vspace{-4mm}
    \label{fig:framework}
\end{figure*}

\subsection{Method}
Figure~\ref{fig:framework} illustrates the \ours\ algorithm. We detail the steps below.

\paragraph{LLM Prompting.}
We first sample examples from the labeled dataset $\mathcal{D}^{\text{labeled}}$ and then few-shot prompt an LLM with those to produce pseudolabels for unlabeled data $\mathcal{D}^{\text{unlabeled}}$.

\paragraph{Intermediate KD stages.}\looseness=-1
Since we focus on tasks where few-shot prompted LLMs exhibit room for improvement, we consider the possibility that models finetuned on a sufficient amount of data, even if pseudolabeled by an LLM teacher or a model distilled from it, might be able to learn the task better. In Section~\ref{sec:fidelity}, we will show empirical evidence that this conjecture is indeed true and analyze this phenomenon. We will present the discovery that, during KD from a few-shot prompted LLM or an intermediate model distilled from the LLM, (1) the student performance improves and approaches 100\% F1 on teacher pseudolabels, and (2), meanwhile, the student performance on held-out evaluation data improves and surpasses that of the teacher.

To leverage the above discovery in \ours, we perform data partitioning and propose \textbf{cross-partition labeling} with a collaborative pair of student models at each intermediate KD stage. 
Concretely, we partition unlabeled data $\mathcal{D}^{\text{unlabeled}}$ randomly and evenly into two distillation sets $\mathcal{D}^{\text{unlabeled}}_A$ and $\mathcal{D}^{\text{unlabeled}}_B$. 
Then, at each intermediate KD stage, we \textbf{(1)} train a collaborative pair of students using the mutually exclusive data partitions with pseudolabels generated by model(s) from the previous stage, which can be the LLM teacher or a pair of distilled previous-stage students, and \textbf{(2)} let each current-stage student produce pseudolabels for the partition that it has not been trained on, in order to gradually improve the quality of pseudolabels for $\mathcal{D}^{\text{unlabeled}}$ over stages. 

\looseness=-1
\textbf{Why do we need cross-partition labeling?} Cross-partition labeling is key to the operation of \ours.  
Since the student model almost perfectly fits its training data, i.e., pseudolabels produced by the previous-stage model(s), during KD (Section \ref{sec:fidelity}), letting the student label its own training partition would reproduce nearly the same pseudolabels that it was trained on. Using those pseudolabels to train a next-stage student can only be expected to yield very little improvement, if any.
We overcome this problem by partitioning the unlabeled data so that each student is asked to label a partition that it was not trained on.

\paragraph{The final KD stage.}
In the last distillation stage, a single student model is trained on the entire $\mathcal{D}^{\text{unlabeled}}$ with the latest pseudolabels.

\subsection{KD mechanism}
For sequence-level knowledge distillation, the teacher can provide multiple types of supervision signals for the student, such as, token-level logits, greedy beams (hard pseudolabels), and beams that are close to the ground truth. \citet{kim-rush-2016-sequence} have observed that greedy beams are typically the most effective, although the other two may marginally further increase the student performance.
In this work, we adopt greedy beams generated by the teacher for KD, following the literature on KD from LLMs~\citep{wang-etal-2021-want-reduce,yoo-etal-2021-gpt3mix-leveraging,ding2022gpt,li2023feasibility,gilardi2023chatgpt}. \ours\ can thus be used to distill both open-source and closed-source LLMs where only generated beams are provided.

\section{Experimental Setup}\label{sec:setup}
\subsection{Data and evaluation}\label{subsec:data}
\paragraph{Tasks and datasets.}
We perform experiments on two sequence prediction tasks: constituency parsing and task-oriented semantic parsing, where the outputs are structured sequences and a general-purpose LLM pretrained to generate natural language text has to undergo distribution shifts.  
For constituency parsing, we use the Penn Treebank (PTB) dataset~\citep{PennTreeBank} from the news domain and the Colorado Richly Annotated Full-Text (CRAFT) corpus from the biomedical domain~\citep{verspoor2012corpus}.
Semi-supervised learning for biomedical tasks is especially valuable since data annotation usually requires expertise and is formidably expensive.\footnote{For example, it took 80 annotators around 2.5 years to parse the 20k sentences in the CRAFT corpus.}
For task-oriented semantic parsing, we use the Airline Travel Information Systems (ATIS) dataset~\citep{tur2010left} and the Snips dataset.\footnote{\url{https://github.com/snipsco/nlu-benchmark/tree/master/2017-06-custom-intent-engines}}
These datasets have been used for joint intent classification and slot filling.
We focus on the latter that is a sequence generation task, letting models input natural langauge queries and output sequences of per-word slot labels.
More details about the datasets are shown in Appendix~\ref{app:data}.

\paragraph{Simulating low-resource scenarios.}
For the constituency parsing datasets, PTB and CRAFT, we randomly sample 250 examples from the training set as the pool of labeled data. In different experiments, we randomly sample 50 or 250 examples from this pool to be the labeled set $\mathcal{D}^\text{labeled}$. 
For the task-oriented semantic parsing datasets, ATIS and Snips, we randomly sample 50 examples as the labeled set $\mathcal{D}^\text{labeled}$ since the datasets are smaller. 

We use the remaining training examples, excluding the 250 above for constituency parsing and 50 for semantic parsing, as the pool of unlabeled data.
For the main results (Section \ref{sec:results}),
we randomly sample 20k unlabeled examples for PTB and use the entire unlabeled pool for CRAFT, ATIS, and Snips as the set $\mathcal{D}^\text{unlabeled}$. 
We randomly and evenly partition $\mathcal{D}^\text{unlabeled}$ into $\mathcal{D}^\text{unlabeled}_A$ and $\mathcal{D}^\text{unlabeled}_B$. 
For further analyses (Section \ref{sec:analysis}), the amount of unlabeled data randomly sampled from the pool will be described in the respective contexts.


\paragraph{Evaluation.} 
We compute token-level F1-score for each example and average over examples to measure model performance. 
For constituency parsing, we follow common practice to use bracketing but discard constituent tags in training and evaluation, and remove punctuation before computing parsing F1.\footnote{Decisions associated with punctuation are often arbitrary in ground truth parse trees and not informative of model performance.} 
We minimally post-process model outputs, balancing brackets and fixing accidental word segmentation.\footnote{We do not constrain model generation. Early during training, T5 fails to consistently output well-formed parse trees. This is less of an issue after the first few epochs.}
For semantic parsing, we use the evaluation function from \citet{goo2018slot}.\footnote{\url{https://github.com/MiuLab/SlotGated-SLU/blob/master/utils.py}} 


\subsection{Model, learning, and prediction}\label{subsec:model-learning-prediction}
\paragraph{The LLM teacher.}  We use GPT-3.5 Turbo\footnote{\url{https://platform.openai.com/docs/models}}, a model supported by the ChatGPT API, to collect initial pseudolabels for $\mathcal{D}^\text{unlabeled}$. We access the model through the OpenAI API. We randomly and uniformly sample 30 labeled parsing examples from $\mathcal{D}^\text{labeled}$ to include in the prompt, since 50 or 250 examples will exceed the allowed maximum context length. Details of prompt design can be found in Appendix~\ref{app:implementations}.

\paragraph{The student model.}\looseness=-1
We finetune T5-Base (220M parameters; \citealp{raffel2020t5}) as the student model using the Huggingface transformers library \cite{wolf-etal-2020-transformers}. 
Our assumed low-resource semi-supervised setting implies that the amount of available labeled data is small and a reliably large validation set does not exist. 
Inspired by classic unsupervised algorithms such as K-Means, we train each student until a convergence criteria is met --- we train the student until the training accuracy on teacher pseudolabels changes by less than 0.1\% for three consecutive epochs. 
The other hyperparameters are learning rate $3\times10^{-4}$, batch size 32, and $21$ maximum training epochs.\footnote{Oracle experiments on PTB show that the validation performance doesn't change significantly given a wide range of these hyperparameters, and thus we use a default set in all experiments. 
In scenarios where a labeled validation set is available, further tuning may prove even more beneficial due to optimization behavior such as double descent \cite{Opper1989BasinsOA,Cun1991EigenvaluesOC,Maddox2020RethinkingPC}.} 
Generation is performed via greedy decoding. 
Results are from single runs that take less than 6 hours on a single RTX 6000 GPU unless otherwise stated.


\subsection{Baselines}\looseness=-1
We compare \ours\ with the few-shot prompted LLM, T5-Base finetuned using few-shot labeled data $\mathcal{D}^\text{labeled}$, and the following KD baselines.

\paragraph{Vanilla KD.}
We employ sequence-level knowledge distillation~\citep{kim-rush-2016-sequence} where a student is trained on hard labels generated by GPT-3.5 Turbo for the unlabeled dataset $\mathcal{D}^\text{unlabeled}$.

\paragraph{KD + SD w/ filtering.}
We first train a student with teacher pseudolabels on the entire unlabeled dataset $\mathcal{D}^{\text{unlabeled}}$ through vanilla KD. 
Then we apply self-distillation (SD): letting the student predict pseudolabels for $\mathcal{D}^{\text{unlabeled}}$, filtering them by keeping $r\%$ of the data where the student has the highest average token log-probabilities, and retraining a student on the filtered pseudolabeled data. 
Confidence-based filtering is a common method to optimize learning from pseudolabels~\citep{lang2022co,wang2021selective,mohananey-etal-2020-self,vinyals2015grammar,DBLP:conf/naacl/McCloskyCJ06}. We note that KD + SD without filtering is similar to vanilla KD, since the student pseudolabels on $\mathcal{D}^{\text{unlabeled}}$ can be expected to be almost identical with teacher pseudolabels. We do not report a baseline that similarly filters LLM pseudolabels before we train a student, since the GPT-3.5 Turbo API access does not provide logits. 

\begin{table*}[!t]
\centering
\small
\begin{tabular}{l|cc|cc|c|c}
\toprule
 \textbf{Dataset} & \multicolumn{2}{c|}{\textbf{PTB}}  & \multicolumn{2}{c|}{\textbf{CRAFT}}  & \multicolumn{1}{c|}{\textbf{ATIS}}  & \multicolumn{1}{c}{\textbf{Snips}}\\
 \textbf{$N_{\text{labeled}}$} & \textbf{50} &\textbf{250} &\textbf{50} &\textbf{250} &\textbf{50} &\textbf{50}\\
\midrule
GPT-3.5 Turbo & 69.1 & 70.7 & 60.0 & 59.9 & 84.6 &87.9\\
T5-Base & 0 & 59.7 & 0 & 54.1 & 34.1 &30.6 \\
\midrule
Vanilla KD & 71.6 & 74.7 & 63.8 & 63.7 &89.5 &89.3\\
KD + SD &72.1  &74.7 &64.3 &63.8 &89.9 & 89.5\\
\multicolumn{1}{r|}{~w/filter: keep top 75\%} &72.5 &75.2 &64.7 &65.3 &89.7 &83.8\\
\multicolumn{1}{r|}{~w/filter: keep top 50\%} &72.4 & 74.6 &64.3 & 64.9 &89.9 &78.8\\
\multicolumn{1}{r|}{~w/filter: keep top 25\%} &73.1 &73.8 &62.2 & 62.9 &87.9 &70.8\\
\midrule
2-stage \ours & {73.9} & {76.6} & {66.9} & {67.5} & 91.4 &90.9\\
3-stage \ours & \textbf{74.2} & \textbf{76.9} & 67.5 & 68.7 & \textbf{91.8} & 91.4\\
4-stage \ours & \textbf{74.2} &\textbf{76.9} &\textbf{68.5} &\textbf{69.3} &\textbf{91.8} &\textbf{91.7} \\
\bottomrule
\end{tabular}
\caption{
The test F1 on PTB, CRAFT, ATIS, and Snips datasets, where $N_\text{labeled}=50$ or $250$ gold annotated sentences are available. GPT-3.5 Turbo is accessed via an API using few-shot prompts; all other settings finetune T5-Base. \ours\ outperforms baselines in all settings.} \vspace{-4mm}
\label{tab:exp1}
\end{table*}

\section{Main Results}\label{sec:results}

\paragraph{Baseline performance.}
As shown in Table~\ref{tab:exp1}, GPT-3.5 Turbo's few-shot performance is reasonably good but has room for improvement. 
Due to the maximum input length for GPT-3.5 Turbo, we always sample 30 in-context demonstrations. 
In most cases, sampling them from a set of 50 or 250 labeled data $\mathcal{D}^\text{labeled}$ does not make a big difference.

Supervised finetuning of T5-Base on $\mathcal{D}^\text{labeled}$ has the worst performance. 
This matches our assumption that the amount of available labeled data is too small to directly finetune a model. 
On both datasets for constituency parsing, 50 labeled examples completely fail to effectively finetune T5-Base to generate well-formed parse trees.

Vanilla KD yields students that outperform the teacher across all datasets. 
KD + SD with confidence-based filtering is shown to have limited performance improvement over vanilla KD and is subject to the selection of threshold for high-confidence labels for each case. For instance, keeping the 25\% most confident pseudolabels gives the best performance for PTB with 50 labeled data, while keeping 75\% performs the best for CRAFT with 50 labeled data. Selecting a proper threshold thus can be resource consuming and difficult in practice especially in low-data regimes with no labeled validation data.

\paragraph{\ours\ performance.} \looseness=-1000
Our \ours\ approach substantially outperforms the few-shot prompted GPT-3.5 Turbo and KD baselines in all settings. For example, on CRAFT with 50 gold labeled training examples, 2-stage \ours\ outperforms GPT-3.5 Turbo by 6.9\% F1 and Vanilla KD by 3.1\% F1. 3-stage \ours\ further improves results, outperforming GPT-3.5 Turbo by 7.5\% and Vanilla KD by 3.7\%. 

3-stage \ours\ performs better than 2-stage \ours\ consistently, but requires additional training time. The 4th-stage KD has marginal benefits in some settings. In practice, 2-stage \ours\ is a good starting point and practitioners can increase the number of stages based on available resources. 

\begin{figure}[t]
    \begin{subfigure}{0.45\textwidth}
        \centering
        \includegraphics[width=.8\linewidth]{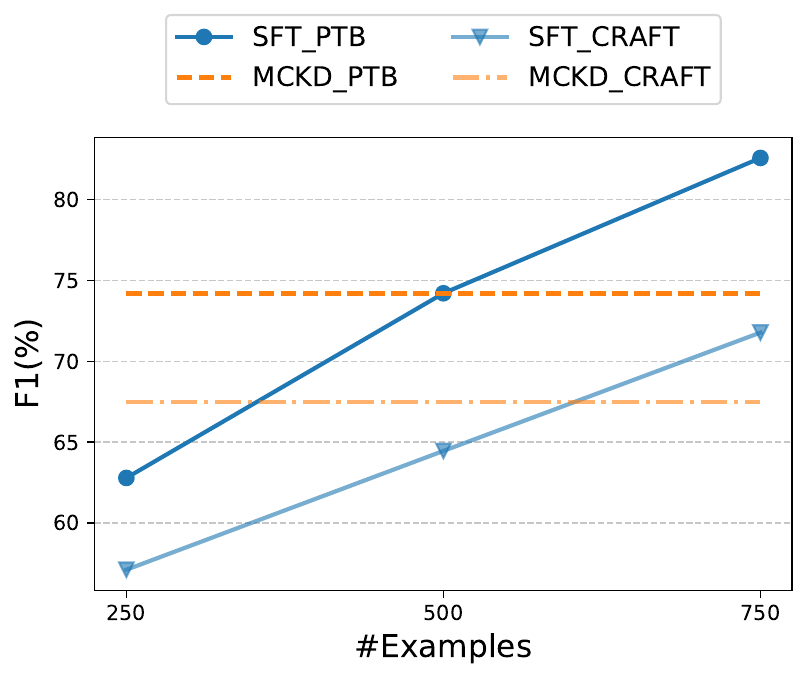}
        \caption{Constituency parsing. The performance of \ours\ using 50 annotations (shown in horizontal lines) is competitive with supervised finetuning (SFT) using 500 or more annotations on PTB and CRAFT.}
    \label{fig:sft-parse}
    \end{subfigure}
    
    \begin{subfigure}{0.45\textwidth}
    \centering
    \includegraphics[width=.8\linewidth]{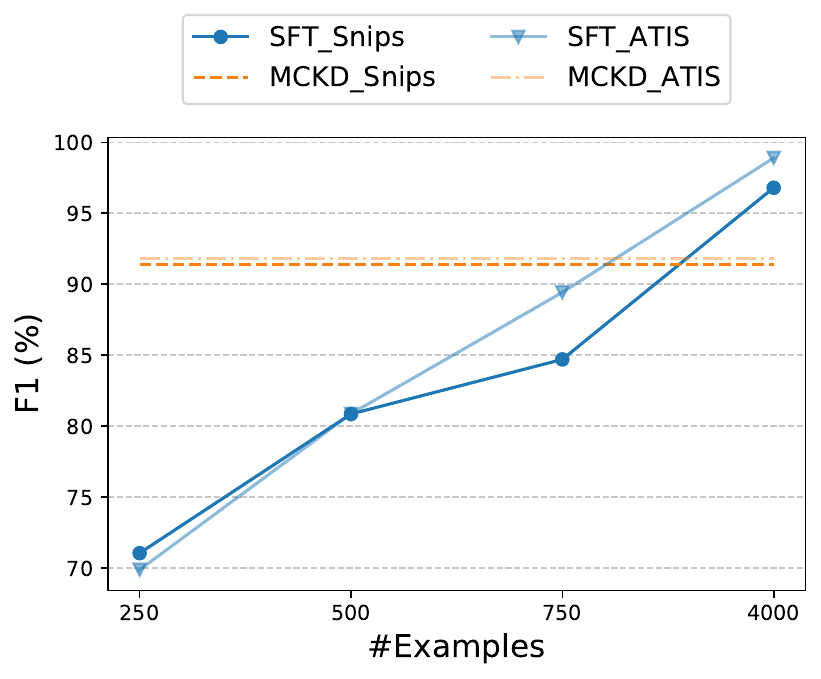}
    \caption{Task-oriented semantic parsing. The performance of \ours\ using 50 annotations (shown in horizontal lines) outperforms supervised finetuning (SFT) with 750 annotations on ATIS and Snips.}
    \label{fig:sft-slot-fill}
    \end{subfigure}
    
    \caption{The test F1 scores of supervised finetuned (SFT) models increase with more gold annotated data. \ours\ needs much less labeled data to match the performance of SFT.}\vspace{-4mm}
    \label{fig:sft}
\end{figure}

\paragraph{Comparison with direct supervised finetuning.}
For constituency parsing, 3-stage \ours\ using 50 labeled examples can match the performance of supervised finetuning (SFT) of T5-Base using 500 labeled examples on PTB, and outperform SFT using 500 labeled examples on CRAFT (Figure~\ref{fig:sft}a). For task-oriented semantic parsing, 3-stage \ours\ with 50 labeled examples can substantially outperform SFT on 750 labeled examples on both ATIS and Snips (Figure~\ref{fig:sft}b). These results demonstrate the label-efficiency of \ours\ and its effectiveness in semi-supervised sequence generation tasks where annotated data are scarce. 


\section{Analysis}\label{sec:analysis}

This section presents additional analyses that motivate the design of \ours{} and explore more deeply into why it performs well.
All analyses are conducted on PTB data.
For simplicity and without loss of generality, we examine 2-stage KD with a single student per stage; the analyses can be generalized to multiple students and more stages which are advantageous in more applied settings.

In each experiment, we sample subsets of $\mathcal{D}^{\text{unlabeled}}_{A}$ and $\mathcal{D}^{\text{unlabeled}}_{B}$ 
(see Section \ref{subsec:data}) and denote them as $\mathcal{D}_{A}$ and $\mathcal{D}_{B}$. 
We vary the sizes of $\mathcal{D}_{A}$ and $\mathcal{D}_{B}$ across experiments to study the effect of the distillation set size on performance.
We train a stage-1 student $s_1$ using $\mathcal{D}_{A}$ with pseudolabels from an LLM teacher $t$ and evaluate it on $\mathcal{D}_{B}$ against ground truth labels
(assumed to be available only for analysis purposes and not during the training in \ours).
Then we train a stage-2 student $s_2$ using $\mathcal{D}_{B}$ with pseudolabels from 
$s_1$ and evaluate it on the test set.

\subsection{A student can generalize better than its teacher}\label{sec:fidelity}\looseness=-1

\begin{figure}[t]
    \centering
    \includegraphics[scale=0.43]{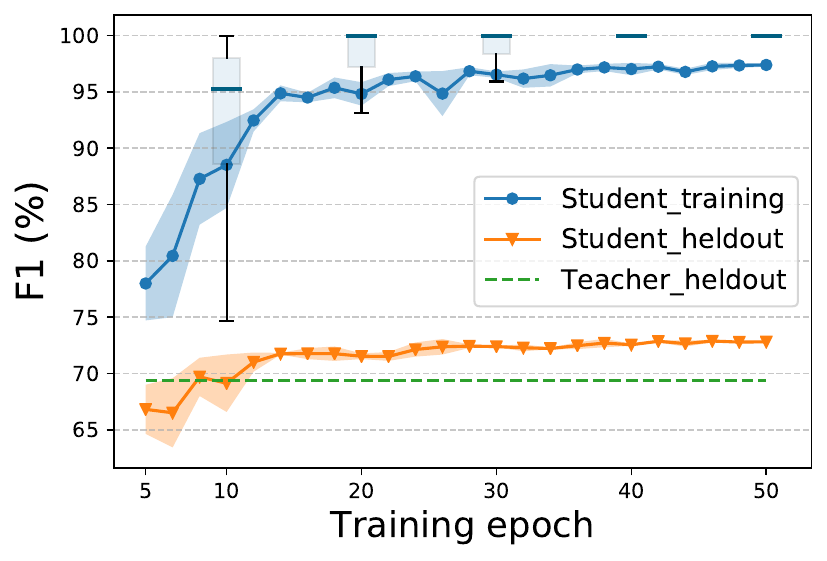}
    \caption{
    The held-out performance of the stage-1 student (F1 against gold annotations) improves 
    with its training performance (F1 against teacher pseudolabels). Its held-out performance also surpasses that of the few-shot prompted LLM teacher.
    }\vspace{-4mm}
    \label{fig:fidelity}
\end{figure}

To understand the generalization properties of a prompted LLM teacher and its KD students, we first train a student $s_1$ on 4k data $\mathcal{D}_{A}$ pseudolabeled by the LLM teacher $t$.
We observe that the student $s_1$ (1)~approaches 100\% F1-score on the pseudolabels of the teacher $t$, and (2)~outperforms the teacher $t$ on 4k held-out data $\mathcal{D}_{B}$ (Figure \ref{fig:fidelity}).
The mean training F1 of $s_1$ on $\mathcal{D}_A$ examples plateaus around 97\%, while the median F1 is 100\% as shown in the box plots, indicating that $s_1$ almost perfectly learns the pseudolabels of $t$, including potentially noisy ones.
However, this does not weaken the generalization of $s_1$, as its performance continues to improve on $\mathcal{D}_{B}$ gold labels, eventually outperforming teacher $t$ by an absolute $\sim13\%$.
This observation holds for various distillation set sizes (see Appendix~\ref{sec:app_data_efficiency_sft}).

To also examine if student $s_1$ can provide better distillation signal than the LLM teacher $t$, we perform the following experiment:
First, we train a stage-2 student $s_2^B$ on 4k data $\mathcal{D}_B$ pseudolabeled by $s_1$. 
We then train a similar student $s_1^B$ on $\mathcal{D}_B$ pseudolabeled by $t$.
Finally, we evaluate $s_1^B$ and $s_2^B$ on the unseen test set.
Our results show that the F1-score of $s_2^B$ (72.5\%) surpasses that of $s_1^B$  (71.3\%), confirming that the pseudolabels of student $s_1$ can train a better model than the LLM psuedolabels. 


%


\begin{figure}
    \centering
    \includegraphics[scale=0.42]{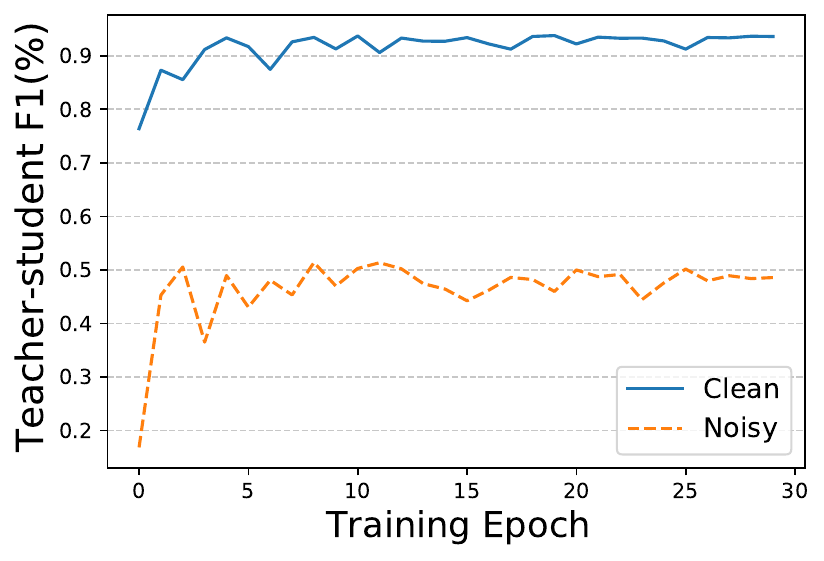}
    \caption{
    The student achieves a high F1 on held-out \textit{clean} pseudolabels from an LLM teacher, but a low F1 on held-out \textit{noisy} pseudolabels from the same teacher.
    The student tends to learn a general distribution over clean teacher pseudolabels while learning noisy ones via instance-wise memorization.
    }\vspace{-4mm}
    \label{fig:mem-gen-label}
\end{figure}

\paragraph{Memorization vs. generalization. }\looseness=-1

Why is the generalization of our student $s_1$ robust to the presence of noisy teacher pseudolabels?
\citet{feldman2020does} shows that the signal in a training example can either be internalized by a trained model as an instance of the general distribution that it learns or memorized separately. 
To find out if $s_1$ learns high and low-quality pseudolabels of teacher $t$ differently in our context, we design the following experiment:
We first identify the examples in the distillation set $\mathcal{D}_A$ where $s_1$ achieves a $100\%$ F1 on pseudolabels of teacher $t$, \ie{}, its training labels.
From these well-learned training examples, we randomly sample 100 \textit{clean} ones on which pseudolabels of $t$ have a 96.3\% average F1 with respect to the ground truth labels, and denote them as $\mathcal{D}_\text{clean}$. 
We also sample 100 \textit{noisy} examples where pseudolabels of $t$ have only a 31.8\% average F1 with respect to ground truth, and denote them as $\mathcal{D}_\text{noisy}$.
We then finetune a new T5-base student $s'_1$ on the remaining data $\mathcal{D}_A - (\mathcal{D}_\text{clean} \cup \mathcal{D}_\text{noisy})$ and evaluate it on $\mathcal{D}_\text{clean}$ and $\mathcal{D}_\text{noisy}$ against pseudolabels of teacher $t$.

As shown in Figure~\ref{fig:mem-gen-label}, $s'_1$ obtains a high teacher-student F1 on $\mathcal{D}_\text{clean}$ but a low teacher-student F1 on $\mathcal{D}_\text{noisy}$.
The original student $s_1$, with $\mathcal{D}_\text{clean}$ and $\mathcal{D}_\text{noisy}$ in its training data, was able to learn both sets well with a 97\% teacher-student F1 in Figure \ref{fig:fidelity}.
When both are removed from training (of $s'_1$), $\mathcal{D}_\text{clean}$ still resembles the general distribution learned by $s'_1$ from the remaining training data, while $\mathcal{D}_\text{noisy}$ is now out-of-distribution.
The $s'_1$ results above thus confirm that our stage-1 \ours{} student learns a general distribution over clean teacher pseudolabels while memorizing the noisy ones in a more instance-specific way. 
This explains why the generalization of the student is not affected by its learning of teacher errors in Figure \ref{fig:fidelity}.

\begin{figure}[t]
    \centering
    \includegraphics[scale=0.42]{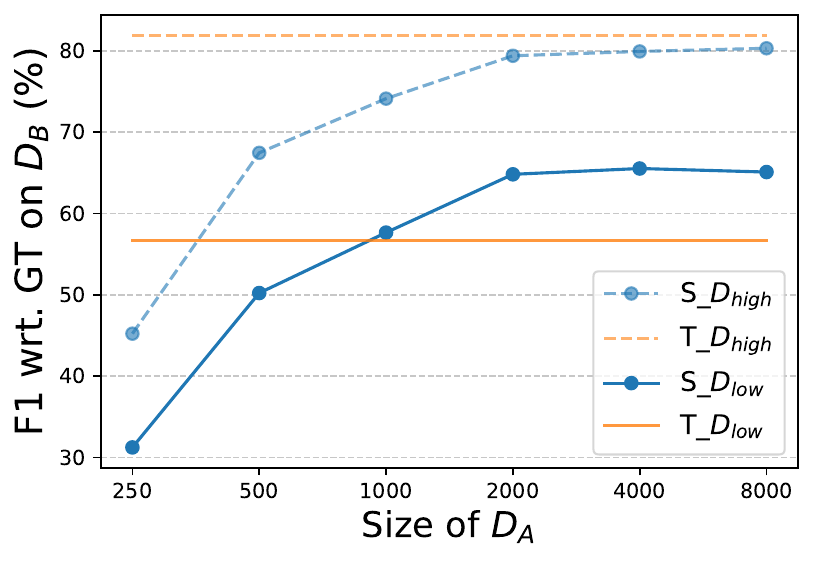}
    \caption{
    Performance (F1) on subsets of the held-out set $\mathcal{D}_{B}$ on which the teacher performs the best ($\mathcal{D}_{high}$) and the worst ($\mathcal{D}_{low}$).
    With sufficient amounts of pseudolabeled data, the student exhibits high agreement with the teacher on $\mathcal{D}_{high}$ while outperforming on $\mathcal{D}_{low}$.
    }\vspace{-4mm}
    \label{fig:heldout_cmp}
\end{figure}

\subsection{The student corrects the teacher's errors}\label{sec:correct} \looseness=-1000
We have observed the student $s_1$ trained on 4k instances of $\mathcal{D}_{A}$ pseudolabeled by the LLM teacher $t$ to outperform its teacher $t$ on the held-out set $\mathcal{D}_{B}$.
To understand on which held-out examples student $s_1$ is outperforming the teacher $t$, we split $\mathcal{D}_{B}$ evenly into two halves: the sentences on which $t$ achieves the highest F1 ($\mathcal{D}_{high}$) and those on which it has the lowest F1 ($\mathcal{D}_{low}$).
As illustrated in Figure~\ref{fig:heldout_cmp}, the performance of $s_1$ improves on both halves given more pseudolabeled training data $\mathcal{D}_{A}$.
Given a large enough $\mathcal{D}_{A}$, the performance of $s_1$ approaches that of $t$ on $\mathcal{D}_{high}$ and is better on $\mathcal{D}_{low}$.
In other words, the student is able to correct the teacher's mistakes while almost completely retaining its true knowledge.

\subsection{Large distillation sets are useful}\label{sec:size}
\begin{figure}[t]
\centering
\includegraphics[scale=0.42]{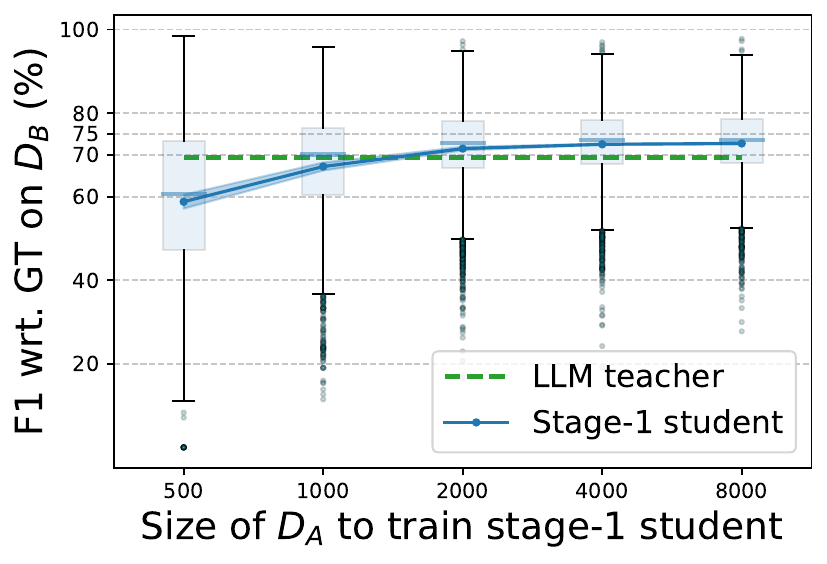}
\caption{
Performance (F1) of the stage-1 student on held-out set $\mathcal{D}_{B}$ when trained on varying amounts of pseudolabeled data from $\mathcal{D}_A$.
With a sufficiently large distillation set, the student's F1 surpasses that of the teacher. 
We observe further improvements with multistage distillation (not shown in the figure).
}\vspace{-4mm}
\label{fig:size_heldout}
\end{figure}

Given an equal amount of labeled data $\mathcal{D}^\text{labeled}$, we investigate in this section if the stage-2 \ours{} student $s_2$ can benefit from growing amounts of unlabeled data $\mathcal{D}^\text{unlabeled}$.
First, Figure~\ref{fig:size_heldout} shows that the performance of the stage-1 student $s_1$ on the held-out set $\mathcal{D}_{B}$ of 4k instances improves as the distillation set $\mathcal{D}_{A}$ grows larger.
With a $\mathcal{D}_{A}$ of 2k or more instances, $s_1$ outperforms the few-shot prompted LLM teacher $t$ on $\mathcal{D}_{B}$.

\begin{figure}[t]
    \centering
    \includegraphics[scale=0.55]{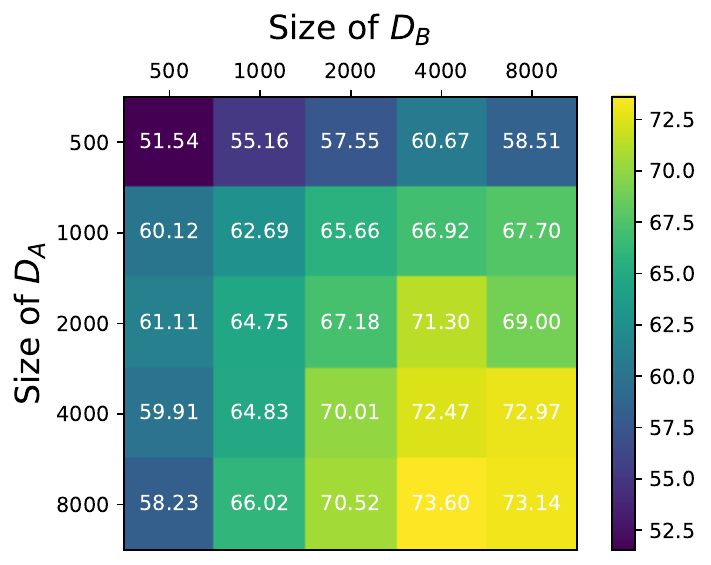}
    \caption{
    Test F1 against ground truth labels with varying amounts of stage-1 and stage-2 pseudolabeled data.
    The pseudolabels used in stage 1 ($\mathcal{D}_{A}$) are inferred by the teacher, and those used in stage 2 ($\mathcal{D}_{B}$) are inferred by the stage-1 student.
    It is generally better to use more data in both stages for stronger final performance, although performance can drop if the difference between the sizes of $\mathcal{D}_{A}$ and $\mathcal{D}_{B}$ is too large.
    }\vspace{-4mm}
    \label{fig:table-ablation-ab}
\end{figure}

A stage-2 student $s_2$ is then trained on $\mathcal{D}_{B}$ with pseudolabels from $s_1$ and evaluated on the test set.
We examine the effect of the sizes of distillation sets $\mathcal{D}_{A}$ and $\mathcal{D}_{B}$ on the final test performance of $s_2$ in  Figure~\ref{fig:table-ablation-ab}.
When the distillation set $\mathcal{D}_{A}$ for $s_1$ is too small, increasing the size of $\mathcal{D}_{B}$ by generating more pseudolabels with $s_1$ does not substantially improve performance for $s_2$, likely due to the pseudolabels being too noisy.
Learning from an increasing amount of such noisy labels may even worsen the test set performance of $s_2$; for instance, with a $\mathcal{D}_{A}$ of 2k instances, enlarging $\mathcal{D}_{B}$ from 4k to 8k instances reduces the test F1 of $s_2$ from 71.3 to 69.0.
On the other hand, when $\mathcal{D}_{B}$ is too small, making $\mathcal{D}_{A}$ larger alone does not improve the test set performance of $s_2$ either, despite higher-quality pseudolabels in $\mathcal{D}_{B}$ from a better-trained $s_1$.
Overall, the above results indicate that adequately large distillation sets are needed at every stage of \ours{} to optimize the performance of the final student.
In the original, more streamlined implementation in Section \ref{sec:mistkd}, we employ a pair of students at each stage to pseudolabel examples, so that the final student can take advantage of the most data available.

\section{Related Work}\label{app:related_work} \looseness=-1
Knowledge distillation (KD), apart from being used for model compression~\cite{tang2019distilling,jiao-etal-2020-tinybert,Bucila2006ModelC,sun-etal-2020-mobilebert}, has shown effectiveness in semi-supervised learning~\cite{NEURIPS2022_2e343555}, where a teacher model annotates unlabeled data to train a student model. 
We study low-resource semi-supervised sequence generation with KD. 
We use hard pseudolabels from teachers to train student models, following common practice in sequence-level knowledge distillation~\cite{kim-rush-2016-sequence}, including scenarios when the teacher is an LLM~\cite{ding2022gpt,yoo-etal-2021-gpt3mix-leveraging,wang-etal-2021-want-reduce,gilardi2023chatgpt,shridhar2022distilling,ho2022large,li2023feasibility}. 

Self-distillation (SD) through training the model with its self-generated pseudolabels is often valuable in enhancing generalization, but finding reliable confidence thresholds to filter the pseudolabels and ensure their quality is often tricky~\cite{Furlanello2018BornAN,liu-etal-2021-noisy}. Instead, we improve the pseudolabel quality over stages by cross-partition labeling, taking advantage of the better generalization abilities of models over stages.

Some past works also leverage several student models to improve the distillation performance. \citet{zhang2018deep} and \citet{guo2020online} introduce several student models that learn from each other's predicted logits. \citet{song2018collaborative} and \citet{zhu2018knowledge} introduce extra branches of the student model (e.g., high-level layers or the classifier heads) to gather their multiple views as additional information for student training, which is shown helpful to improve the generalization. These works are mainly evaluated on image classification and do not analyze distillation from LLMs.
 
Co-training~\cite{Blum1998CombiningLA,han2018co,lang2022co} is relevant to our work since it involves using separate models to generate improved pseudolabels. 
In co-training, different learners are typically initialized with different views of the data. 
Then they provide confident predictions to each other for additional training. 
\citet{wei2021theoretical} provides theoretical explanation on why generated pseudolables can be more credible than original labels for training. 
In contrast, our approach begins with an LLM teacher and trains better student models over stages. 

\section{Conclusion}
Semi-supervised sequence generation in specialized low-resource settings can be challenging for NLP systems.
We present the discovery that student models distilled from a few-shot prompted LLM can often correct the teacher’s errors, thus showing superior generalization, on such tasks.
Exploiting this phenomenon along with strategies such as data partitioning and cross-partition labeling over multiple stages, we then propose \ours{}, a novel solution that utilizes knowledge distillation (KD) from an LLM, whose performance can be only at a mediocre level, to train much smaller yet high-accuracy students.
\ours{} essentially leverages progressively better models to train future generations with even better performance.
Extensive empirical evaluation verifies the effectiveness and label efficiency of \ours{}.

\section{Limitations}
We have studied the scenario of semi-supervised sequence generation where labeled data are too scarce to train a model while a few-shot prompted LLM exhibits room for improvement. 
We therefore use an LLM as the initial teacher for knowledge distillation (KD) and follow common practice in sequence-level KD by using hard labels for KD.
Future work may explore the effectiveness of the proposed multistage KD with cross-partition labeling (1) from a non-LLM teacher and (2) using soft logits or other mechanisms for KD.

The GPT-3.5 Turbo teacher model in our experiments is a company hosted service and many details are missing about how it was implemented and which data it was trained on. 
Relatively more information is available about T5-based. For instance, it was pretrained on large amounts of news data; this may explain why its performance on PTB is relatively higher than that on CRAFT.
\bibliography{custom}

\begin{thebibliography}{44}
\expandafter\ifx\csname natexlab\endcsname\relax\def\natexlab#1{#1}\fi

\bibitem[{Amini et~al.(2022)Amini, Feofanov, Pauletto, Devijver, and Maximov}]{amini2022self}
Massih-Reza Amini, Vasilii Feofanov, Loic Pauletto, Emilie Devijver, and Yury Maximov. 2022.
\newblock Self-training: A survey.
\newblock \emph{arXiv preprint arXiv:2202.12040}.

\bibitem[{Blum and Mitchell(1998)}]{Blum1998CombiningLA}
Avrim Blum and Tom~M. Mitchell. 1998.
\newblock Combining labeled and unlabeled data with co-training.
\newblock In \emph{COLT' 98}.

\bibitem[{Brown et~al.(2020)Brown, Mann, Ryder, Subbiah, Kaplan, Dhariwal, Neelakantan, Shyam, Sastry, Askell, Agarwal, Herbert{-}Voss, Krueger, Henighan, Child, Ramesh, Ziegler, Wu, Winter, Hesse, Chen, Sigler, Litwin, Gray, Chess, Clark, Berner, McCandlish, Radford, Sutskever, and Amodei}]{DBLP:conf/nips/BrownMRSKDNSSAA20}
Tom~B. Brown, Benjamin Mann, Nick Ryder, Melanie Subbiah, Jared Kaplan, Prafulla Dhariwal, Arvind Neelakantan, Pranav Shyam, Girish Sastry, Amanda Askell, Sandhini Agarwal, Ariel Herbert{-}Voss, Gretchen Krueger, Tom Henighan, Rewon Child, Aditya Ramesh, Daniel~M. Ziegler, Jeffrey Wu, Clemens Winter, Christopher Hesse, Mark Chen, Eric Sigler, Mateusz Litwin, Scott Gray, Benjamin Chess, Jack Clark, Christopher Berner, Sam McCandlish, Alec Radford, Ilya Sutskever, and Dario Amodei. 2020.
\newblock Language models are few-shot learners.
\newblock In \emph{Advances in Neural Information Processing Systems 33: Annual Conference on Neural Information Processing Systems 2020, NeurIPS 2020, December 6-12, 2020, virtual}.

\bibitem[{Bucila et~al.(2006)Bucila, Caruana, and Niculescu-Mizil}]{Bucila2006ModelC}
Cristian Bucila, Rich Caruana, and Alexandru Niculescu-Mizil. 2006.
\newblock Model compression.
\newblock In \emph{Knowledge Discovery and Data Mining}.

\bibitem[{Ding et~al.(2022)Ding, Qin, Liu, Bing, Joty, and Li}]{ding2022gpt}
Bosheng Ding, Chengwei Qin, Linlin Liu, Lidong Bing, Shafiq Joty, and Boyang Li. 2022.
\newblock Is gpt-3 a good data annotator?
\newblock \emph{arXiv preprint arXiv:2212.10450}.

\bibitem[{Feldman(2020)}]{feldman2020does}
Vitaly Feldman. 2020.
\newblock Does learning require memorization? a short tale about a long tail.
\newblock In \emph{Proceedings of the 52nd Annual ACM SIGACT Symposium on Theory of Computing}, pages 954--959.

\bibitem[{Furlanello et~al.(2018)Furlanello, Lipton, Tschannen, Itti, and Anandkumar}]{Furlanello2018BornAN}
Tommaso Furlanello, Zachary~Chase Lipton, Michael Tschannen, Laurent Itti, and Anima Anandkumar. 2018.
\newblock Born again neural networks.
\newblock In \emph{International Conference on Machine Learning}.

\bibitem[{Gilardi et~al.(2023)Gilardi, Alizadeh, and Kubli}]{gilardi2023chatgpt}
Fabrizio Gilardi, Meysam Alizadeh, and Ma{\"e}l Kubli. 2023.
\newblock Chatgpt outperforms crowd-workers for text-annotation tasks.
\newblock \emph{arXiv preprint arXiv:2303.15056}.

\bibitem[{Goo et~al.(2018)Goo, Gao, Hsu, Huo, Chen, Hsu, and Chen}]{goo2018slot}
Chih-Wen Goo, Guang Gao, Yun-Kai Hsu, Chih-Li Huo, Tsung-Chieh Chen, Keng-Wei Hsu, and Yun-Nung Chen. 2018.
\newblock Slot-gated modeling for joint slot filling and intent prediction.
\newblock In \emph{Proceedings of the 2018 Conference of the North American Chapter of the Association for Computational Linguistics: Human Language Technologies, Volume 2 (Short Papers)}, pages 753--757.

\bibitem[{Guo et~al.(2020)Guo, Wang, Wu, Yu, Liang, Hu, and Luo}]{guo2020online}
Qiushan Guo, Xinjiang Wang, Yichao Wu, Zhipeng Yu, Ding Liang, Xiaolin Hu, and Ping Luo. 2020.
\newblock Online knowledge distillation via collaborative learning.
\newblock In \emph{Proceedings of the IEEE/CVF Conference on Computer Vision and Pattern Recognition}, pages 11020--11029.

\bibitem[{Han et~al.(2018)Han, Yao, Yu, Niu, Xu, Hu, Tsang, and Sugiyama}]{han2018co}
Bo~Han, Quanming Yao, Xingrui Yu, Gang Niu, Miao Xu, Weihua Hu, Ivor Tsang, and Masashi Sugiyama. 2018.
\newblock Co-teaching: Robust training of deep neural networks with extremely noisy labels.
\newblock \emph{Advances in neural information processing systems}, 31.

\bibitem[{Ho et~al.(2022)Ho, Schmid, and Yun}]{ho2022large}
Namgyu Ho, Laura Schmid, and Se-Young Yun. 2022.
\newblock Large language models are reasoning teachers.
\newblock \emph{arXiv preprint arXiv:2212.10071}.

\bibitem[{Iliopoulos et~al.(2022)Iliopoulos, Kontonis, Baykal, Menghani, Trinh, and Vee}]{NEURIPS2022_2e343555}
Fotis Iliopoulos, Vasilis Kontonis, Cenk Baykal, Gaurav Menghani, Khoa Trinh, and Erik Vee. 2022.
\newblock Weighted distillation with unlabeled examples.
\newblock In \emph{Advances in Neural Information Processing Systems}, volume~35.

\bibitem[{Jiao et~al.(2020)Jiao, Yin, Shang, Jiang, Chen, Li, Wang, and Liu}]{jiao-etal-2020-tinybert}
Xiaoqi Jiao, Yichun Yin, Lifeng Shang, Xin Jiang, Xiao Chen, Linlin Li, Fang Wang, and Qun Liu. 2020.
\newblock {T}iny{BERT}: Distilling {BERT} for natural language understanding.
\newblock In \emph{Findings of the Association for Computational Linguistics: EMNLP 2020}, pages 4163--4174. Association for Computational Linguistics.

\bibitem[{Kim and Rush(2016)}]{kim-rush-2016-sequence}
Yoon Kim and Alexander~M. Rush. 2016.
\newblock Sequence-level knowledge distillation.
\newblock In \emph{Proceedings of the 2016 Conference on Empirical Methods in Natural Language Processing}, pages 1317--1327. Association for Computational Linguistics.

\bibitem[{Kontonis et~al.(2023)Kontonis, Iliopoulos, Trinh, Baykal, Menghani, and Vee}]{kontonis2023slam}
Vasilis Kontonis, Fotis Iliopoulos, Khoa Trinh, Cenk Baykal, Gaurav Menghani, and Erik Vee. 2023.
\newblock Slam: Student-label mixing for distillation with unlabeled examples.
\newblock In \emph{Thirty-seventh Conference on Neural Information Processing Systems}.

\bibitem[{Kung et~al.(2022)Kung, Cheatham, ChatGPT, Medenilla, Sillos, Leon, Elepa{\~n}o, Madriaga, Aggabao, Diaz-Candido, Maningo, and Tseng}]{Kung2022}
Tiffany~H. Kung, Morgan Cheatham, ChatGPT, Arielle Medenilla, Czarina Sillos, Lorie~De Leon, Camille Elepa{\~n}o, Maria Madriaga, Rimel Aggabao, Giezel Diaz-Candido, James Maningo, and Victor Tseng. 2022.
\newblock \href {https://doi.org/10.1101/2022.12.19.22283643} {Performance of chatgpt on usmle: Potential for ai-assisted medical education using large language models}.
\newblock \emph{medRxiv}.

\bibitem[{Lang et~al.(2022)Lang, Agrawal, Kim, and Sontag}]{lang2022co}
Hunter Lang, Monica~N Agrawal, Yoon Kim, and David Sontag. 2022.
\newblock Co-training improves prompt-based learning for large language models.
\newblock In \emph{International Conference on Machine Learning}, pages 11985--12003. PMLR.

\bibitem[{LeCun et~al.(1991)LeCun, Kanter, and Solla}]{Cun1991EigenvaluesOC}
Yann LeCun, Ido Kanter, and Sara~A. Solla. 1991.
\newblock \href {https://api.semanticscholar.org/CorpusID:41596547} {Eigenvalues of covariance matrices: Application to neural-network learning.}
\newblock \emph{Physical review letters}, 66 18:2396--2399.

\bibitem[{Li et~al.(2023)Li, Wang, Ma, Liu, Wang, Wu, and Gao}]{li2023feasibility}
Zongjie Li, Chaozheng Wang, Pingchuan Ma, Chaowei Liu, Shuai Wang, Daoyuan Wu, and Cuiyun Gao. 2023.
\newblock On the feasibility of specialized ability stealing for large language code models.
\newblock \emph{arXiv preprint arXiv:2303.03012}.

\bibitem[{Liu et~al.(2021)Liu, Shen, and Lapata}]{liu-etal-2021-noisy}
Yang Liu, Sheng Shen, and Mirella Lapata. 2021.
\newblock Noisy self-knowledge distillation for text summarization.
\newblock In \emph{Proceedings of the 2021 Conference of the North American Chapter of the Association for Computational Linguistics: Human Language Technologies}, pages 692--703. Association for Computational Linguistics.

\bibitem[{Maddox et~al.(2020)Maddox, Benton, and Wilson}]{Maddox2020RethinkingPC}
Wesley~J. Maddox, Gregory~W. Benton, and Andrew~Gordon Wilson. 2020.
\newblock \href {https://api.semanticscholar.org/CorpusID:211989398} {Rethinking parameter counting in deep models: Effective dimensionality revisited}.
\newblock \emph{ArXiv}, abs/2003.02139.

\bibitem[{Marcus et~al.(1993)Marcus, Santorini, and Marcinkiewicz}]{PennTreeBank}
Mitchell~P. Marcus, Beatrice Santorini, and Mary~Ann Marcinkiewicz. 1993.
\newblock Building a large annotated corpus of english: The penn treebank.
\newblock \emph{Comput. Linguistics}, 19(2):313--330.

\bibitem[{McClosky et~al.(2006)McClosky, Charniak, and Johnson}]{DBLP:conf/naacl/McCloskyCJ06}
David McClosky, Eugene Charniak, and Mark Johnson. 2006.
\newblock Effective self-training for parsing.
\newblock In \emph{Human Language Technology Conference of the North American Chapter of the Association of Computational Linguistics, Proceedings, June 4-9, 2006}. The Association for Computational Linguistics.

\bibitem[{McClosky et~al.(2008)McClosky, Charniak, and Johnson}]{DBLP:conf/coling/McCloskyCJ08}
David McClosky, Eugene Charniak, and Mark Johnson. 2008.
\newblock When is self-training effective for parsing?
\newblock In \emph{{COLING} 2008, 22nd International Conference on Computational Linguistics, Proceedings of the Conference, 18-22 August 2008}, pages 561--568.

\bibitem[{Mohananey et~al.(2020)Mohananey, Kann, and Bowman}]{mohananey-etal-2020-self}
Anhad Mohananey, Katharina Kann, and Samuel~R. Bowman. 2020.
\newblock Self-training for unsupervised parsing with {PRPN}.
\newblock In \emph{Proceedings of the 16th International Conference on Parsing Technologies and the IWPT 2020 Shared Task on Parsing into Enhanced Universal Dependencies}, pages 105--110. Association for Computational Linguistics.

\bibitem[{Opper et~al.(1989)Opper, Kleinz, K{\"o}hler, and Kinzel}]{Opper1989BasinsOA}
Manfred Opper, J.~Kleinz, H.~K{\"o}hler, and Wolfgang Kinzel. 1989.
\newblock \href {https://api.semanticscholar.org/CorpusID:119540013} {Basins of attraction near the critical storage capacity for neural networks with constant stabilities}.
\newblock \emph{Journal of Physics A}, 22.

\bibitem[{Raffel et~al.(2020)Raffel, Shazeer, Roberts, Lee, Narang, Matena, Zhou, Li, and Liu}]{raffel2020t5}
Colin Raffel, Noam Shazeer, Adam Roberts, Katherine Lee, Sharan Narang, Michael Matena, Yanqi Zhou, Wei Li, and Peter~J. Liu. 2020.
\newblock Exploring the limits of transfer learning with a unified text-to-text transformer.
\newblock \emph{J. Mach. Learn. Res.}, 21(1).

\bibitem[{Shridhar et~al.(2022)Shridhar, Stolfo, and Sachan}]{shridhar2022distilling}
Kumar Shridhar, Alessandro Stolfo, and Mrinmaya Sachan. 2022.
\newblock Distilling multi-step reasoning capabilities of large language models into smaller models via semantic decompositions.
\newblock \emph{arXiv preprint arXiv:2212.00193}.

\bibitem[{Singhal et~al.(2022)Singhal, Azizi, Tu, Mahdavi, Wei, Chung, Scales, Tanwani, Cole-Lewis, Pfohl et~al.}]{Singhal2022LargeLM}
Karan Singhal, Shekoofeh Azizi, Tao Tu, S~Sara Mahdavi, Jason Wei, Hyung~Won Chung, Nathan Scales, Ajay Tanwani, Heather Cole-Lewis, Stephen Pfohl, et~al. 2022.
\newblock Large language models encode clinical knowledge.
\newblock \emph{arXiv preprint arXiv:2212.13138}.

\bibitem[{Song and Chai(2018)}]{song2018collaborative}
Guocong Song and Wei Chai. 2018.
\newblock Collaborative learning for deep neural networks.
\newblock \emph{Advances in neural information processing systems}, 31.

\bibitem[{Sun et~al.(2020)Sun, Yu, Song, Liu, Yang, and Zhou}]{sun-etal-2020-mobilebert}
Zhiqing Sun, Hongkun Yu, Xiaodan Song, Renjie Liu, Yiming Yang, and Denny Zhou. 2020.
\newblock {M}obile{BERT}: a compact task-agnostic {BERT} for resource-limited devices.
\newblock In \emph{Proceedings of the 58th Annual Meeting of the Association for Computational Linguistics}, pages 2158--2170. Association for Computational Linguistics.

\bibitem[{Tang et~al.(2019)Tang, Lu, Liu, Mou, Vechtomova, and Lin}]{tang2019distilling}
Raphael Tang, Yao Lu, Linqing Liu, Lili Mou, Olga Vechtomova, and Jimmy Lin. 2019.
\newblock Distilling task-specific knowledge from bert into simple neural networks.
\newblock \emph{arXiv preprint arXiv:1903.12136}.

\bibitem[{Tur et~al.(2010)Tur, Hakkani-T{\"u}r, and Heck}]{tur2010left}
Gokhan Tur, Dilek Hakkani-T{\"u}r, and Larry Heck. 2010.
\newblock What is left to be understood in atis?
\newblock In \emph{2010 IEEE Spoken Language Technology Workshop}, pages 19--24. IEEE.

\bibitem[{Verspoor et~al.(2012)Verspoor, Cohen, Lanfranchi, Warner, Johnson, Roeder, Choi, Funk, Malenkiy, Eckert et~al.}]{verspoor2012corpus}
Karin Verspoor, Kevin~Bretonnel Cohen, Arrick Lanfranchi, Colin Warner, Helen~L Johnson, Christophe Roeder, Jinho~D Choi, Christopher Funk, Yuriy Malenkiy, Miriam Eckert, et~al. 2012.
\newblock A corpus of full-text journal articles is a robust evaluation tool for revealing differences in performance of biomedical natural language processing tools.
\newblock \emph{BMC bioinformatics}, 13(1):1--26.

\bibitem[{Vinyals et~al.(2015)Vinyals, Kaiser, Koo, Petrov, Sutskever, and Hinton}]{vinyals2015grammar}
Oriol Vinyals, {\L}ukasz Kaiser, Terry Koo, Slav Petrov, Ilya Sutskever, and Geoffrey Hinton. 2015.
\newblock Grammar as a foreign language.
\newblock \emph{Advances in neural information processing systems}, 28.

\bibitem[{Wang et~al.(2021{\natexlab{a}})Wang, Yan, Meng, and Zhou}]{wang2021selective}
Fusheng Wang, Jianhao Yan, Fandong Meng, and Jie Zhou. 2021{\natexlab{a}}.
\newblock Selective knowledge distillation for neural machine translation.
\newblock In \emph{Proceedings of the 59th Annual Meeting of the Association for Computational Linguistics and the 11th International Joint Conference on Natural Language Processing (Volume 1: Long Papers)}, pages 6456--6466.

\bibitem[{Wang et~al.(2023)Wang, Zhou, and Sachan}]{wang-etal-2023-lets}
Ruida Wang, Wangchunshu Zhou, and Mrinmaya Sachan. 2023.
\newblock Let{'}s synthesize step by step: Iterative dataset synthesis with large language models by extrapolating errors from small models.
\newblock In \emph{Findings of the Association for Computational Linguistics: EMNLP 2023}, pages 11817--11831, Singapore. Association for Computational Linguistics.

\bibitem[{Wang et~al.(2021{\natexlab{b}})Wang, Liu, Xu, Zhu, and Zeng}]{wang-etal-2021-want-reduce}
Shuohang Wang, Yang Liu, Yichong Xu, Chenguang Zhu, and Michael Zeng. 2021{\natexlab{b}}.
\newblock Want to reduce labeling cost? {GPT}-3 can help.
\newblock In \emph{Findings of the Association for Computational Linguistics: EMNLP 2021}. Association for Computational Linguistics.

\bibitem[{Wei et~al.(2021)Wei, Shen, Chen, and Ma}]{wei2021theoretical}
Colin Wei, Kendrick Shen, Yining Chen, and Tengyu Ma. 2021.
\newblock \href {https://openreview.net/forum?id=rC8sJ4i6kaH} {Theoretical analysis of self-training with deep networks on unlabeled data}.
\newblock In \emph{International Conference on Learning Representations}.

\bibitem[{Wolf et~al.(2020)Wolf, Debut, Sanh, Chaumond, Delangue, Moi, Cistac, Rault, Louf, Funtowicz, Davison, Shleifer, von Platen, Ma, Jernite, Plu, Xu, Le~Scao, Gugger, Drame, Lhoest, and Rush}]{wolf-etal-2020-transformers}
Thomas Wolf, Lysandre Debut, Victor Sanh, Julien Chaumond, Clement Delangue, Anthony Moi, Pierric Cistac, Tim Rault, Remi Louf, Morgan Funtowicz, Joe Davison, Sam Shleifer, Patrick von Platen, Clara Ma, Yacine Jernite, Julien Plu, Canwen Xu, Teven Le~Scao, Sylvain Gugger, Mariama Drame, Quentin Lhoest, and Alexander Rush. 2020.
\newblock \href {https://doi.org/10.18653/v1/2020.emnlp-demos.6} {Transformers: State-of-the-art natural language processing}.
\newblock In \emph{Proceedings of the 2020 Conference on Empirical Methods in Natural Language Processing: System Demonstrations}, pages 38--45, Online. Association for Computational Linguistics.

\bibitem[{Yoo et~al.(2021)Yoo, Park, Kang, Lee, and Park}]{yoo-etal-2021-gpt3mix-leveraging}
Kang~Min Yoo, Dongju Park, Jaewook Kang, Sang-Woo Lee, and Woomyoung Park. 2021.
\newblock {GPT}3{M}ix: Leveraging large-scale language models for text augmentation.
\newblock In \emph{Findings of the Association for Computational Linguistics: EMNLP 2021}. Association for Computational Linguistics.

\bibitem[{Zhang et~al.(2018)Zhang, Xiang, Hospedales, and Lu}]{zhang2018deep}
Ying Zhang, Tao Xiang, Timothy~M Hospedales, and Huchuan Lu. 2018.
\newblock Deep mutual learning.
\newblock In \emph{Proceedings of the IEEE conference on computer vision and pattern recognition}, pages 4320--4328.

\bibitem[{Zhu et~al.(2018)Zhu, Gong et~al.}]{zhu2018knowledge}
Xiatian Zhu, Shaogang Gong, et~al. 2018.
\newblock Knowledge distillation by on-the-fly native ensemble.
\newblock \emph{Advances in neural information processing systems}, 31.

\end{thebibliography}
\bibliographystyle{acl_natbib}

\clearpage

\appendix

\section{Dataset Details}\label{app:data}
In Table \ref{tab:dt_example}, we show examples of the contituency and semantic parsing tasks that we work with. We use the training and testing sets described in Section \ref{subsec:data}. We assume a low-resource scenario where a labeled validation set is unavailable; we find default parameters that are not tuned on dataset-specific validation sets are effective across datasets.
For PTB we adopt the standard data splits.


\paragraph{Constituency parsing.}
The standard split for the test set of PTB contains 2,416 examples.  For training, we randomly sample 20,000 examples as unlabeled training data. CRAFT contains 21,121 examples, but does not have standard splits; we randomly sample 50\% examples as the training set and 40\% examples as the test set. The remaining could be used for validation, although we assume a low-resource scenario where a labeled validation set is unavailable. For CRAFT, the test set contains 8,448 examples, while the unlabeled training set contains 10,560 examples.

\paragraph{Semantic parsing.}
For ATIS, the training set contains 4,478 utterances and the test set contains 893 utterances. There are 120 slot labels, i.e., labels for words. For Snips, the training set contains 13,084 utterances and the test set contains 700 utterances. There are 72 slot labels. We adopt dataset splits by~\citet{goo2018slot}.

\begin{table*}[t]
\centering
\makebox[\textwidth]{
\begin{tabular}{l}
\toprule
\textbf{Constituency parsing on PTB}\\
\textbf{Input}: Some lousy earnings reports whacked the stock market but bond prices fell only slightly \\and the dollar rose a little against most major currencies.\\
\textbf{Output}: ( ( ( ( Some ) ( lousy ) ( earnings ) ( reports ) ) ( ( whacked ) ( ( the ) ( stock ) ( market ) ) ) )\\ ( but ) ( ( ( bond ) ( prices ) ) ( ( fell ) ( ( only ) ( slightly ) ) ) ) ( and ) ( ( ( the ) ( dollar ) ) ( ( rose )\\ ( ( a ) ( little ) ) ( ( against ) ( ( most ) ( major ) ( currencies ) ) ) ) ) )\\
\\
\textbf{Constituency parsing on CRAFT}\\
\textbf{Input}: Rather , the abnormal organization of Sertoli cells appears * to result from lack of \\Dmrt7 in the germ line.\\
\textbf{Output}: ((((Rather)) (,) (((the) (abnormal) (organization)) ((of) ((Sertoli) (cells)))) \\((appears) (((*)) ((to) ((result) ((from) (((lack)) ((of) ((Dmrt7))) ((in) ((the) (germ) (line))))))))) (.)) )\\
\\
\textbf{Semantic parsing on ATIS}\\
\textbf{Input}: list the fares of US Air flights from Boston to Philadelphia \\
\textbf{Output}: O O O O B-airline\_name B-airline\_name O O B-fromloc.city\_name O B-toloc.city\_name\\
\\
\textbf{Semantic parsing on SNIPS}\\
\textbf{Input}: add sabrina salerno to the grime instrumentals playlist \\
\textbf{Output}: O B-artist I-artist O O B-playlist I-playlist O\\
\bottomrule
\end{tabular}}
\caption{Examples from our used datasets.}
\label{tab:dt_example}
\end{table*}

\section{Prompting an LLM Teacher for Pseudolabel Generation}\label{app:implementations}
We employ the standard random selection approach~\citep{DBLP:conf/nips/BrownMRSKDNSSAA20} to choose in-context exemplars from available labeled data to prompt the LLM. The prompt format is shown in Table~\ref{tab:prompt}.

For task-oriented semantic parsing, we use the following instruction for ATIS and Snips respectively: ``Predict tag for each word in the sentence. `O' for unrelated word. Focus on words related to the flight information, such as locations, time and airline.''; ``Predict tag for each word in the sentence. `O' for unrelated word. Focus on words related to the music, restuarant, weather, book, playlist and searching.''  We also preprocess the output of in-context exemplars to interleave each predicted tag with its according slot word so as to improve the prompting performance for LLMs. For instance, for the example shown in Table~\ref{tab:dt_example}, the output when prompting LLM is formatted as ``list O the O fares O of O US B-airline\_name Air B-airline\_name flights from O Boston B-fromloc.city\_name to O Philadelphia B-toloc.city\_name''.

\begin{table}[t]
\centering
\begin{tabular}{l}
\toprule
...\\
\textbf{Input}:\{input string of exemplar $i$\}\\
\textbf{Output}:\{output result of exemplar $i$\}\\
...\\
\textbf{Input}:\{input string of test data\}\\
\textbf{Output}:\\
\bottomrule
\end{tabular}
\caption{The prompting format for LLM.}
\label{tab:prompt}
\end{table}

\section{Statistical Significance of Results}
We present the statistical significance of the results in Table \ref{tab:exp1}.
On each dataset, for the setting that uses 50 labeled examples, we perform a paired t-test to compare (1) the distribution of per-example F1 scores of predictions by the best KD baseline and (2) that of predictions by the 2-stage \ours. Under the null hypothesis, the distribution of the per-example F1 scores achieved by the strongest baseline model is the same as the per-example F1 scores achieved by the 2-stage MCKD. In Table~\ref{tab:app-p-value}, we observe that the p-values are all far below the commonly used 0.05 threshold. Therefore, we reject the null hypotheses and conclude that the 2-stage \ours\ is significantly better than the strongest baselines in the mentioned dataset settings.

\begin{table*}[t]
\centering
\begin{tabular}{lcccc}
\hline
\textbf{Dataset} & \textbf{PTB}     & \textbf{CRAFT}   & \textbf{ATIS} & \textbf{Snips} \\ \hline
Best baseline    & KD+SD (top 25\%) & KD+SD (top 75\%) & KD+SD         & KD+SD          \\
p-value          & 5.7e-10          & 5.3e-22          & 9.9e-5        & 1.2e-3         \\ \hline
\end{tabular}
\caption{P-values of paired t-tests between the per-example F1 distribution of our 2-stage \ours\ and that of the best baselines. We observe that the \ours\ is significantly better than the other approaches across different datasets.}
\label{tab:app-p-value}
\end{table*}

\section{Relation Between Training Fidelity and Generalization Ability}
\label{sec:app_data_efficiency_sft}
\begin{figure}[t]
    \centering
\includegraphics[scale=0.45]{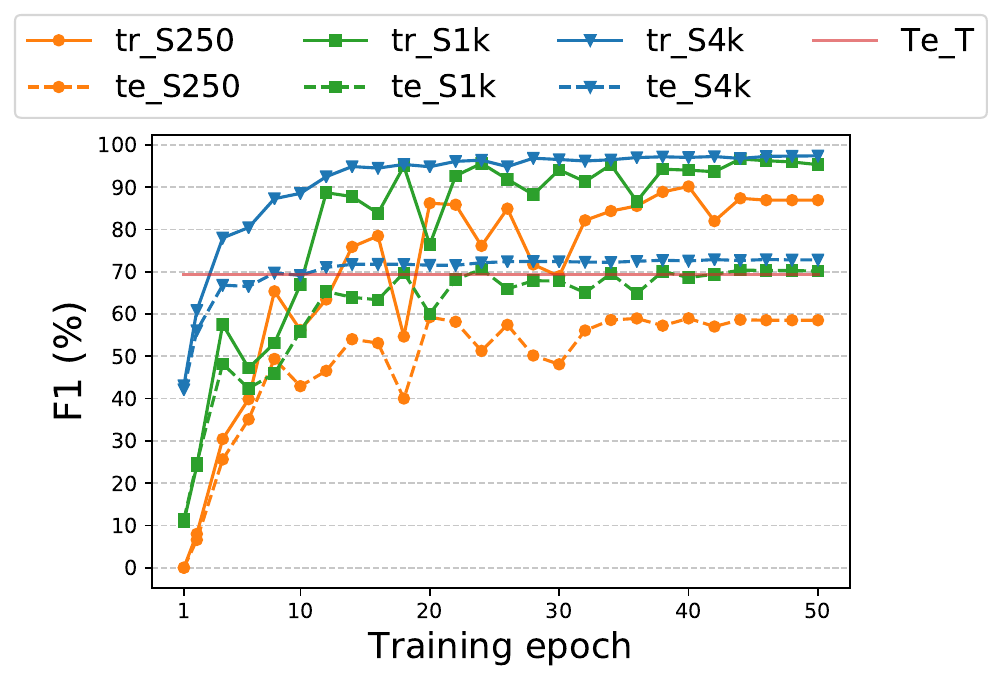}
    \caption{Student training F1 on teacher pseudolabels and student evaluation F1 on held-out data with ground truth labels using different distillation set sizes. E.g., ``tr\_S250'' represents the training F1 of a student trained with 250 teacher pseudolabels, while ``te\_S250'' is its generalization performance on held-out data. The generalization performance continues to improve when the student better learns teacher pseudolabels during training. }
    \label{fig:training_fidelity_size}
\end{figure}

As discussed in Section~\ref{sec:fidelity}, we find high training fidelity of a student model is generally correlated with better performance on held-out data, despite the unreliability of teacher labels. This section investigates whether the observation still holds with different scales of the distillation set, especially when the size is substantially smaller. Results are shown in Figure~\ref{fig:training_fidelity_size}. Similar patterns across different scales of distillation sets can be observed. Even if the distillation set only contains 250 teacher pseudolabels, the observation holds true.

\end{document}